%% file: main.tex
\documentclass[letterpaper,10pt,times,mathptm,psfig,conference]{IEEEtran}
\IEEEoverridecommandlockouts
\usepackage[letterpaper, left=19mm, right=19mm, top=19mm, bottom=19mm]{geometry}
\usepackage{cite}
\usepackage{amsmath,amssymb,amsfonts}
\usepackage{algorithmic}
\usepackage[ruled,linesnumbered]{algorithm2e}
\usepackage[normalem]{ulem}
\usepackage{amssymb}
\usepackage{amsmath}
\usepackage{graphicx}
\usepackage{textcomp}
\usepackage{url}
\usepackage{multirow}
\usepackage{booktabs}
\usepackage{array}
\usepackage{tabularx}
\usepackage{xcolor}
\usepackage{pifont}
\usepackage{booktabs,multirow,makecell}

\usepackage{epsfig, amsmath, amssymb, wrapfig}
\usepackage{subcaption}
\def\BibTeX{{\rm B\kern-.05em{\sc i\kern-.025em b}\kern-.08em
    T\kern-.1667em\lower.7ex\hbox{E}\kern-.125emX}}

\newcommand{\khalil}[1]{}
\author{
Md Khalid Syfullah and Alvi Ataur Khalil\\
Transformative Innovation for Trustworthy AI and Network Security (TITANS) Lab, \\ Computer Science, Southern Illinois University Carbondale, USA\\
\{mdkhalid.syfullah, a.khalil\}@siu.edu
\vspace{-10pt}}

\begin{document}
\title{LLM-Based Schema-Aware Split Learning for Privacy-Preserving Mental Distress Prediction Across Heterogeneous Surveys
\vspace{-10pt}}

\maketitle
\thispagestyle{empty}
\pagestyle{empty}

\begin{abstract}
\input{Sections/abstract}
\end{abstract}


\input{Sections/introduction}

\input{Sections/background}

\input{Sections/related_works}

\input{Sections/methodology}

\input{Sections/experimental_result}

\input{Sections/conclusion}


\bibliographystyle{IEEEtran}
\bibliography{References}

\end{document}

%% file: Sections/abstract.tex
Rising societal and lifestyle complexity has been linked to a growing prevalence of mental distress worldwide. Educational institutions, workplaces, clinics, etc. collect large volumes of mental health survey data to understand and reduce this burden. Collaborative analysis of such data could yield effective generalizable predictive models. Privacy constraints and varied survey designs (i.e., different questions, scales, and formats) hinder direct integration. We propose a schema-aware split learning (SL) framework that preserves privacy, using a large language model (LLM) as a shared semantic encoder to harmonize heterogeneous survey schemas across institutions. We serialize each survey record into a natural-language description, unifying disparate survey schemas into a common format. The LLM is fine-tuned for mental distress assessment via Low-Rank Adaptation (LoRA) and partitioned across client and server. Clients retain the raw survey responses locally and run only a lightweight front-end, so original records never leave the institution that collected them. The resource-intensive backbone runs on the server, minimizing client-side computation. Using LLaMA-3.2-3B-Instruct, the framework attains an average ANLS of 0.708 with only 2,000 training samples, surpasses federated learning (FL) in eight of nine settings, and cuts per-client computation by three orders of magnitude, while generalizing to unseen datasets. Overall, it enables accurate, privacy-preserving, and resource-efficient collaborative learning from heterogeneous mental health survey data.

%% file: Sections/introduction.tex
\section{Introduction}
\label{sec:introduction}

Mental disorders are among today's most pressing public health challenges. The World Health Organization (WHO) estimates that in 2022, roughly one in eight people worldwide (nearly one billion individuals) were living with a mental disorder, including anxiety or depression, with the burden falling more heavily on students and working-age populations~\cite{world2022world, cuijpers2023world}. Stress and related conditions are often assessed using self-report instruments, including the Depression, Anxiety and Stress Scales (DASS)~\cite{lovibond1995structure}. Different organizations conduct mental health surveys every year~\cite{samhsa2023nsduh}. These surveys support automated mental distress screening with predictive AI models~\cite{shatte2019machine}.


Despite these advances in AI models, two main challenges limit the collaborative use of these data. First, they are highly sensitive, and unauthorized disclosure can cause stigma and discrimination~\cite{thornicroft2022lancet}. In addition, data-protection regulations such as HIPAA and the GDPR restrict the sharing of health records, keeping datasets siloed within individual institutions and permitting their use only under strict privacy safeguards~\cite{hipaa1996,voigt2017eu}. Second, institutions use different survey instruments, response scales, and question formats, creating inconsistent feature spaces and data distributions, a problem known as data heterogeneity~\cite{gao2022survey}. An effective AI-based assessment system must therefore address privacy protection and cross-schema heterogeneity simultaneously. However, existing methods typically tackle only one of these issues, and, to the best of our knowledge, no prior work jointly addresses both for mental distress assessment.


A Large Language Model (LLM) can act as a semantic encoder to address schema heterogeneity in conventional models~\cite{ye2024towards}. When a survey record is converted into a natural-language description, for example, ``sleep duration: less than 5 hours'' rather than a coded value like ``sleep: 3'', the meaning of each feature is expressed through language rather than encoded in a fixed vector position. As a result, surveys that capture similar concepts through different questions, scales, or formats can be mapped into a shared representation, even when their original schemas differ. Recent work on language-based tabular learning confirms this serialization transfers knowledge across datasets~\cite{hegselmann2023tabllm,dinh2022lift,yan2024making}.

Our goal is to let institutions collaboratively train a mental-distress model without sharing raw responses, which we achieve through two components. First, we adapt the LLM with Low-Rank Adaptation (LoRA)~\cite{hu2022lora}, which freezes the pretrained weights and trains only small adapters, sharply reducing trainable parameters and the memory and communication cost of collaborative training. Second, a Split Learning (SL) architecture keeps raw data on the client: each holds only the tokenizer, a LoRA-adapted embedding, and a lightweight projection head, while the LLM backbone and classifier stay on the server. Accordingly, our privacy model is data locality: the survey responses and identifiers never leave the institution that collected them, and only cut-layer activations and their gradients cross the client--server boundary. We do not claim that these intermediate tensors are non-invertible; hardening them against reconstruction (model-inversion) attacks, for instance through differential privacy or activation obfuscation, is outside the scope of this work. To the best of our knowledge, this is the first framework to combine SL for lightweight, data-local clients, a LoRA-adapted LLM as a semantic encoder, and language-based harmonization of heterogeneous survey schemas for mental distress assessment. One of the related works uses a frozen language model with federated averaging instead of SL, forcing every client to run the full model, ignores heterogeneous schemas, and targets cardiac and financial tables rather than mental health surveys~\cite{gaber2025federated}. Recent SL frameworks for LLMs target general NLP tasks and device heterogeneity, not heterogeneous survey schemas~\cite{lin2026hsplitlora}. The contributions of this paper are fourfold:

\begin{itemize}
\item We propose a privacy-preserving SL framework that trains a LoRA-adapted LLaMA-3.2-3B-Instruct~\cite{meta2024llama32} collaboratively across institutional clients and a server, so that raw responses remain local to each institution.
\item We introduce a schema-aware serialization that turns heterogeneous survey records into LLM prompts by mapping ordinal responses to text, marking missing values, and adding a schema header, so surveys with differing feature sets share a common representation.
\item We evaluate the framework under varying data imbalance and schema heterogeneity, benchmarking it against Federated Learning (FL), zero-shot, few-shot, and LoRA baselines and a second LLM OpenBioLLM-8B~\cite{OpenBioLLMs}.
\item We analyze the framework's generalization to unseen survey datasets by training on a small data pool.
\end{itemize}

The study is guided by the following research questions:
\begin{itemize}
\item \textbf{RQ1:} Can an LLM-based SL framework predict mental distress more accurately than conventional baselines when clients hold different survey schemas?
\item \textbf{RQ2:} How does performance change as clients vary in dataset size and survey structure?
\item \textbf{RQ3:} How does SL compare with FL in performance and per-client computational cost?
\end{itemize}

The remainder of this paper is organized as follows. Section~\ref{sec:background} introduces the necessary background, Section~\ref{sec:literature-review} reviews related work, Section~\ref{sec:methodology} presents the methodology, Section~\ref{sec:experimental} reports the results, and Section~\ref{sec:conclusion} concludes.

%% file: Sections/background.tex
\section{Background}
\label{sec:background}

This section introduces the concepts behind our framework: LLM, LLMs and FT-Transformer for survey data, LoRA, SL.

\subsection{LLMs, LLaMA and OpenBioLLM}
\label{sec:bg_llama}
LLMs build on the Transformer architecture~\cite{vaswani2017attention} and learn semantic representations from large text corpora, making them useful beyond traditional NLP. In this work, we use LLaMA-3.2-3B-Instruct~\cite{meta2024llama32}, a compact model from Meta's LLaMA family of open-weight LLMs that balances performance and efficiency~\cite{grattafiori2024llama}. We also evaluate OpenBioLLM-8B~\cite{OpenBioLLMs}, a biomedical model built on LLaMA-3-8B and fine-tuned on curated clinical and medical corpora, which reports state-of-the-art results on clinical NLP benchmarks.

\subsection{Survey Data with LLMs and FT-Transformer}
\label{sec:bg_survey}
Recent studies show tabular and survey data can be serialized into natural language and processed effectively by LLMs~\cite{dinh2022lift,yan2024making}. This maps related questions into a shared representation even when institutions use different features, scales, or coding schemes. Beyond LLMs, FT-Transformer~\cite{gorishniy2021revisiting} tokenizes each tabular feature and uses attention to model feature interactions, achieving strong performance on structured data. However, it learns a feature space directly from the data rather than feature semantics, limiting transfer across datasets with differing schemas~\cite{hegselmann2023tabllm}.

\subsection{Low-Rank Adaptation (LoRA)}
\label{sec:bg_lora}
Fine-tuning all LLM parameters is computationally expensive and memory intensive. LoRA~\cite{hu2022lora} keeps the pretrained weights fixed and trains only a small set of low-rank adapters, preserving most benefits of full fine-tuning. This suits adapting large models to domain-specific tasks with limited resources.
\subsection{Split Learning (SL)}
\label{sec:bg_sl}
\begin{figure}[t]
\centering
\includegraphics[width=0.9\columnwidth]{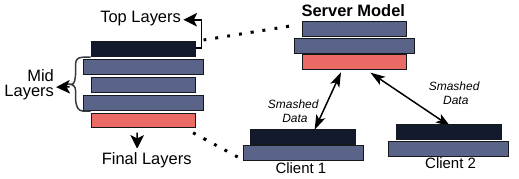}
\vspace{-5pt}
\caption{Overview of the split-learning framework.}
\vspace{-15pt}
\label{fig:back}
\end{figure}

SL divides a neural network between clients and a server, keeping data local while exchanging intermediate representations during training~\cite{vepakomma2018split}. This improves privacy and allows resource-constrained clients to participate without hosting the model. Fig.~\ref{fig:back} shows an overview of SL framework.

%% file: Sections/related_works.tex
\section{Literature Review}
\label{sec:literature-review}

Both SL and FL enable collaboration without sharing raw data, differing in how they distribute computation: SL partitions the model across client and server, whereas FL shares the model. In SL, clients process data and share intermediate activations~\cite{vepakomma2018split}, and SplitFed improves scalability through parallel training~\cite{thapa2022splitfed}. FL provides an alternative, with FedAvg~\cite{mcmahan2017communication} as the most widely used scheme.

\begin{figure*}[t]
\centering
\includegraphics[width=0.8\linewidth]{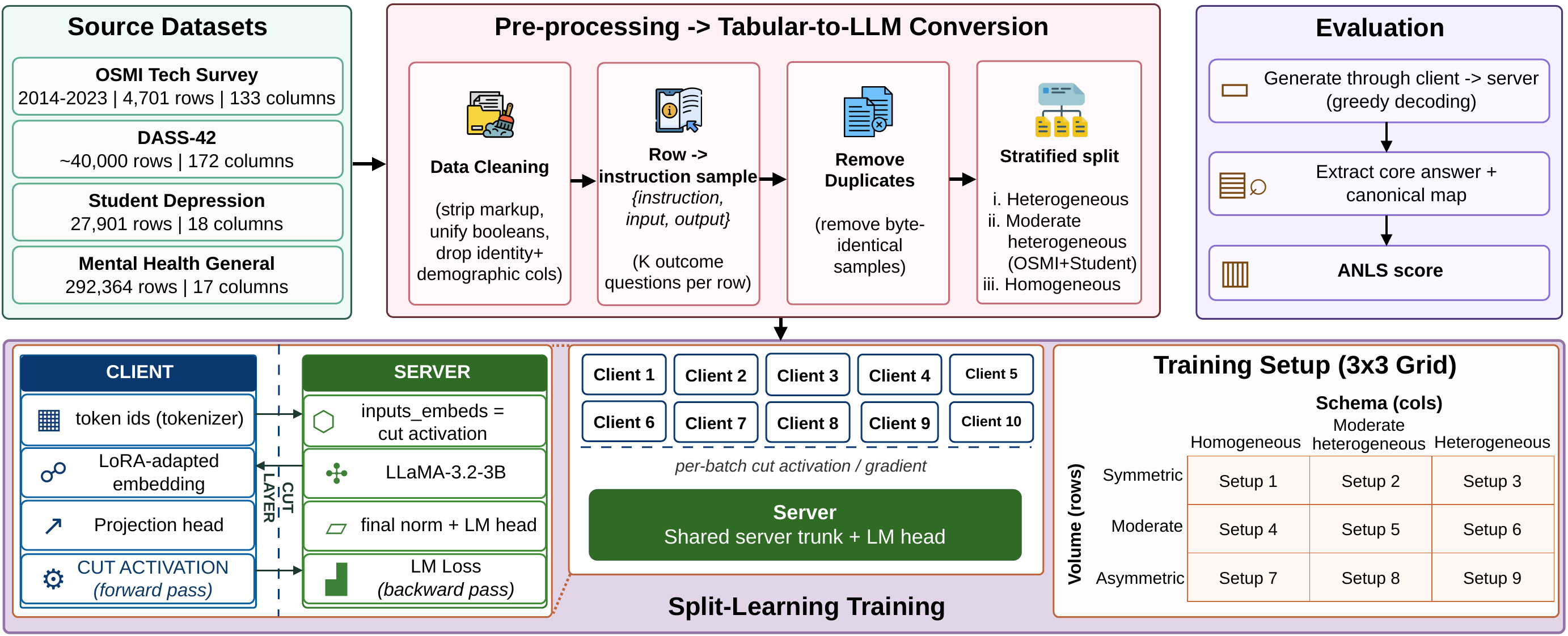}
\vspace{-5pt}
\caption{System overview of the proposed split-learning framework.}
\vspace{-15pt}
\label{fig:overview}
\end{figure*}

Parameter-Efficient Fine-Tuning (PEFT) has made LLMs practical in distributed settings: LoRA~\cite{hu2022lora} trains only a few parameters, while SplitLoRA~\cite{lin2024splitlora} and HSplitLoRA~\cite{lin2026hsplitlora} extend this to SL and FL. These approaches reduce communication and computation, letting clients join LLM-backbone training without sharing raw data~\cite{lin2024splitlora}.

A separate line of work uses LLMs as semantic encoders for tabular data by serializing records into text. TabLLM~\cite{hegselmann2023tabllm}, LIFT~\cite{dinh2022lift}, and TP-BERTa~\cite{yan2024making} show that semantically related features from different schemas can share a representation space, while FT-Transformer~\cite{gorishniy2021revisiting} provides a strong non-LLM baseline for structured data. FedLLM-Align~\cite{gaber2025federated} brings this to heterogeneous tables in a federated setting, yet uses a frozen encoder with federated averaging and targets non-clinical data. These studies confirm the value of semantic serialization, yet none addresses a central need: enabling resource-constrained institutions to learn collaboratively from heterogeneous mental health surveys without exposing sensitive responses. SL targets this by keeping raw data local and offloading the heavy LLM backbone to the server. In mental health, machine learning has predicted depression and stress from survey data, with privacy-preserving extensions through FL~\cite{khalil2024federated}, yet these works assume a common feature schema.

Across these areas, existing methods satisfy at most two of the three needs in our setting: SL for lightweight, data-local clients; a LoRA-adapted LLM encoder; and support for heterogeneous survey schemas~\cite{lin2024splitlora,lin2026hsplitlora,hegselmann2023tabllm,dinh2022lift,gaber2025federated}. None combines all three for mental distress prediction, and none demonstrates strong cross-schema performance from a small training pool. Our framework fills this gap by unifying schema-aware survey serialization, a LoRA-based semantic encoder, and SL into a privacy-preserving solution that is lightweight and data-efficient on the client.
\vspace{-5pt}
\begin{table}[hbt!]
\centering
\caption{List of Notations.}
\vspace{-5pt}
\label{tab:notation}
\scriptsize
\renewcommand{\arraystretch}{1.15}
\begin{tabular}{|c|l|}
\hline
Symbol & Description \\
\hline
$N$            & Number of SL clients \\
\hline
$\mathbf{x}$   & Serialized survey record (prompt) \\
\hline
$\mathbf{a}$   & Cut-layer activation sent client$\rightarrow$server \\
\hline
$\mathbf{g}$   & Gradient returned server$\rightarrow$client \\
\hline
$\mathbf{W}_0$ & Frozen pretrained weight matrix \\
\hline
$\mathbf{A},\mathbf{B}$ & LoRA factors ($\Delta\mathbf{W}=\tfrac{\alpha}{r}\mathbf{B}\mathbf{A}$) \\
\hline
$r,\alpha$     & LoRA rank and scaling constant \\
\hline
$\tau$         & ANLS non-answer threshold \\
\hline
$p,g$          & Predicted and ground-truth core answers \\
\hline
\end{tabular}
\vspace{-10pt}
\end{table}

%% file: Sections/methodology.tex
\section{Methodology}
\label{sec:methodology}

This section presents the framework: schema-aware serialization, data partitions, the SL architecture, and evaluated LLM paradigms. Fig.~\ref{fig:overview} shows the system overview.

\subsection{Schema-Aware Serialization}
\label{sec:serialization}
\label{sec:data}
We use four public mental health surveys spanning domains, schemas, and label spaces: the OSMI Mental Health in Tech Survey (OST)~\cite{osmi2014}, DASS-42 (D-42)~\cite{dass42_openpsychometrics}, the Student Depression Dataset (SD)~\cite{student_depression_kaggle}, and a general-population Mental Health dataset (MHD)~\cite{mental_health_general_kaggle}. A unified pipeline cleans each dataset, harmonizes column names, unifies boolean encodings, and applies DASS validity checks. Identifiable and demographic columns are removed before model building.

\begin{table*}[t]
\centering
\caption{Record-to-prompt serialization across the four datasets.}
\label{tab:serialization}
\vspace{-5pt}
\scriptsize
\renewcommand{\arraystretch}{1.3}
\begin{tabular}{|p{0.04\textwidth}|c|p{0.42\textwidth}|p{0.15\textwidth}|p{0.18\textwidth}|}
\hline
\textbf{Dataset} & \textbf{\#Outcomes} & \textbf{Example input (serialized features)} & \textbf{Example instruction} & \textbf{Example output} \\
\hline
OST & 7 &
Employer provides mental health benefits: No. Family history of mental illness: Yes. Ease of taking medical leave: Somewhat difficult. &
Did this tech worker seek professional mental health treatment? &
Yes, this tech worker has sought professional mental health treatment. \\
\hline
D-42 & 7 &
Symptom responses --- Q3 (inability to experience positive feeling): most of the time; \ldots\ Personality --- anxious and easily upset: agree strongly. &
What is this person's depression severity level? &
This person's depression severity is: Severe (score: 24/42). \\
\hline
SD & 5 &
Academic pressure (0--5): 5. Sleep duration: less than 5 hours. Dietary habits: Unhealthy. Financial stress (1--5): 4. &
Does this person show signs of depression? &
Yes, this person is experiencing depression.\\
\hline
MHD & 6 &
Occupation: Corporate. Days spent indoors: more than 2 months. Mood swing frequency: High. Noticed changes in habits: Yes. &
Is this person experiencing growing stress? &
Yes, this person is experiencing growing stress. \\
\hline
\end{tabular}
\vspace{-15pt}
\end{table*}
Each survey record, i.e., one respondent's row in a cleaned dataset, is serialized into multiple instruction-following samples in the \textit{\{instruction, input, output\}} format, one per outcome, so that the model learns each survey's multiple distress-related outcomes instead of a single fixed input-to-label mapping. Table~\ref{tab:serialization} summarizes this mapping. The predicted outcome is removed from the input to prevent leakage, and exact (input, output) duplicates are dropped before any split. Because a single respondent yields several samples, all splitting is performed at the respondent level rather than the sample level: every sample derived from a given respondent is assigned to exactly one of the training, validation, or test sets, so no respondent ever appears in more than one split and no near-duplicate of a training record can leak into evaluation.
\subsection{Data Partitions}
\label{sec:partitions}
Serialization places all surveys in a shared language space, letting datasets with different formats merge into one corpus. A model trained on this mix acquires broader knowledge than one tied to a single schema. To study this, we construct three partitions with increasing schema heterogeneity, all stratified by source with a fixed seed and grouped by respondent:
\begin{itemize}
    \item \textbf{Homogeneous:} All four datasets used separately.
    \item \textbf{Moderate Heterogeneous (MHg):} Merged OST and SD, two surveys that share binary distress outcomes but differ in feature schemas and target populations.
    \item \textbf{Fully Heterogeneous (Hg):} All four datasets merged.
    
\end{itemize}
For evaluation, 500 samples are held out from each of the six test sets across all setups. These held-out samples are drawn from respondents that appear in no training or validation split, and the same respondent-disjoint constraint applies when the training pool is sharded across clients, confining each respondent to a single client.

\subsection{Split-Learning Architecture}
\label{sec:arch}
The framework partitions a LoRA-adapted, decoder-only LLaMA-3.2-3B between clients and a server. LoRA injects trainable rank-$r$ matrices into the projection layers of each transformer block; for $\mathbf{W}_0$, the adapted forward pass is
\begin{equation}
h = \mathbf{W}_0\,x + \frac{\alpha}{r}\,\mathbf{B}\mathbf{A}\,x,
\quad \mathbf{A}\in\mathbb{R}^{r\times n},\ \mathbf{B}\in\mathbb{R}^{m\times r},
\label{eq:lora}
\end{equation}
where only $\mathbf{A}$ and $\mathbf{B}$ are updated, reducing trainable parameters to roughly $0.75\%$ of the model. The model is split as follows, with all symbols defined in Table~\ref{tab:notation}:
\begin{itemize}
    \item \textbf{Client:} Each client holds the tokenizer, a LoRA-adapted embedding, and a two-layer projection head ($\sim$100K parameters). It maps token IDs to a cut activation $\mathbf{a}$ and transmits only $\mathbf{a}$ to the server. The projection head is zero-initialized as a residual MLP, so it begins as an identity mapping, keeps $\mathbf{a}$ within the distribution expected by the trunk, and learns only a correction.
    \item \textbf{Server:} The server holds the full LoRA-adapted transformer trunk and the language-model head. It consumes $\mathbf{a}$ as input embeddings, generates answer tokens, computes the loss against ground truth, and returns the cut gradient $\mathbf{g}$ to complete the client's backward pass.
\end{itemize}

Activations and gradients are exchanged per batch, so only cut-layer tensors cross the client-server boundary; the original survey responses, identifiers, and labels remain on the client by construction and are never transmitted. To study SL under realistic conditions, we form a $3\times3$ grid of nine setups by crossing three schema-heterogeneity levels with three data-volume imbalance levels. Schema heterogeneity ranges from homogeneous (one dataset, IID shards) to moderate (two or three datasets) to high (four datasets, one source per client). Volume imbalance ranges from symmetric (10\% per client) to moderate (power-law split) to high (60\% held by one client).

\subsection{LLM Paradigms}
\label{sec:paradigms}
We evaluate four configurations with increasing supervision and decentralization. All share the system prompt, \textit{``You are a mental health assessment assistant. Answer the question concisely and directly based on the given information''}. The first three serve as baselines:
\begin{itemize}
\item \textbf{Zero-shot:} The model receives only the task instruction and serialized test record, then generates predicted label greedily without examples or parameter updates.
\item \textbf{Few-shot:} The model receives 10 topically matching labeled examples and the serialized test record. Examples with the same answer as the test sample are excluded to avoid label leakage.
\item \textbf{LoRA fine-tuning:} The LoRA adapters from Eq.~\eqref{eq:lora} are trained centrally using the full training set. The loss is computed only on assistant response tokens, so the model learns to generate the target label without being penalized for the input prompt.
\end{itemize}

%% file: Sections/experimental_result.tex
\section{Experimental Results}
\label{sec:experimental}

We evaluate the proposed framework: the setup and metric, centralized baselines, SL results, the FL comparison, cross-dataset generalization, the FT-Transformer reference, and research-question findings.

\subsection{Experimental Setup and Metric}
\label{sec:setup}
Experiments used PyTorch on two NVIDIA RTX 6000 Blackwell GPUs (96~GB each) and 32~GB RAM. The default backbone is LLaMA-3.2-3B-Instruct, with OpenBioLLM-8B for cross-family checks. All LLM configurations use LoRA rank $16$, $\alpha{=}32$ on the $q,k,v,o,\text{gate},\text{up},\text{down}$ projections, a $1024$-token limit, and learning rate $2\times10^{-4}$. Centralized LoRA uses $2{,}000$ training samples, batch $32$, and $3$ epochs. SL and FL both use $10$ clients, batch $8$, and $3$ local epochs over the $3\times3$ grid, with at most $2{,}000$ samples split across clients. The FT-Transformer uses $8$ heads, $3$ layers, batch $256$, learning rate $1\times10^{-4}$, and up to $30$ epochs.

All LLM configurations are scored with Average Normalized Levenshtein Similarity (ANLS), adapted for generative answers. Each raw output is mapped to a canonical core answer via priority-ordered extraction and vocabulary mapping (e.g., ``high risk'' $\rightarrow$ severe), then scored by a thresholded normalized similarity:
\vspace{-5pt}
\begin{equation}
\text{NLS}(p,g)=
\begin{cases}
s & \text{if } s\geq\tau,\\[4pt]
0 & \text{otherwise},
\end{cases}
\qquad s = 1-\dfrac{\text{EditDist}(p,g)}{\max(|p|,|g|)},
\label{eq:nls}
\end{equation}
with $\tau{=}0.5$; adjacent severity levels receive partial credit of $0.5$ for clinical proximity. Dataset-level ANLS is the mean over the 500-sample test set, which contains only respondents absent from the corresponding training and validation splits (Section~\ref{sec:partitions}).

\subsection{Performance under Centralized LLM Baselines}
\label{sec:baselines}
Table~\ref{tab:llm_anls} reports centralized LLM baselines. Zero-shot is weak: LLaMA-3.2-3B averages $0.229$ ANLS, ranging from $0.097$ to $0.336$. Few-shot improves only slightly ($0.279$), showing prompting alone cannot map heterogeneous surveys to distress labels. Per-dataset LoRA, trained on $2{,}000$ samples each, raises the average to $0.665$ ANLS, with best results on D-42 ($0.788$) and OST ($0.761$), establishing LoRA as our framework's main trainable component. OpenBioLLM-8B performs similarly ($0.659$), and Fig.~\ref{fig:backbonegain} shows its gains appear mainly in zero-shot and largely disappear after adaptation, confirming that LLaMA stays competitive despite being almost three times lighter.

\begin{table}[t]
\centering
\caption{Centralized baseline ANLS scores.}
\vspace{-5pt}
\label{tab:llm_anls}
\renewcommand{\arraystretch}{1.15}
\setlength{\tabcolsep}{4pt}
\begin{tabular}{lcccccc}
\toprule
& \multicolumn{2}{c}{Zero-Shot} & \multicolumn{2}{c}{Few-Shot} & \multicolumn{2}{c}{LoRA} \\
\cmidrule(lr){2-3}\cmidrule(lr){4-5}\cmidrule(lr){6-7}
Dataset & Llama & OpenBio & Llama & OpenBio & Llama & OpenBio \\
\midrule
SD    & 0.274 & 0.333 & 0.336 & 0.393 & 0.664 & 0.657 \\
OST   & 0.336 & 0.370 & 0.316 & 0.280 & 0.761 & 0.753 \\
MHg   & 0.302 & 0.314 & 0.353 & 0.375 & 0.631 & 0.650 \\
Hg    & 0.155 & 0.196 & 0.238 & 0.226 & 0.702 & 0.693 \\
MHD   & 0.210 & 0.257 & 0.258 & 0.136 & 0.444 & 0.427 \\
D-42  & 0.097 & 0.155 & 0.170 & 0.164 & 0.788 & 0.776 \\
\bottomrule
\end{tabular}
\vspace{-15pt}
\end{table}

\subsection{SL Performance with Heterogeneous Data}
\label{sec:split}
Table~\ref{tab:sl_fl} reports SL framework's ANLS scores across the $3\times3$ grid alongside FL. SL averages $0.708$ ANLS and stays within $0.660$--$0.759$ across nine settings. Performance peaks under homogeneous schemas at $0.759$ and drops slightly as heterogeneity grows, showing that schema-aware serialization lets clients with different feature sets share one representation. It is stable under volume imbalance: average ANLS shifts by under $0.01$ across the three volume levels.

\begin{table}[htp]
\centering
\caption{SL and FL ANLS across grid settings.}
\label{tab:sl_fl}
\vspace{-5pt}
\renewcommand{\arraystretch}{1.15}
\setlength{\tabcolsep}{4pt}
\scriptsize
\begin{tabular}{lccccccccc}
\toprule
Volume & \multicolumn{3}{c}{Symmetric} & \multicolumn{3}{c}{Moderate} & \multicolumn{3}{c}{High} \\
\cmidrule(lr){2-4}\cmidrule(lr){5-7}\cmidrule(lr){8-10}
Schema & Homo & Mod & High & Homo & Mod & High & Homo & Mod & High \\
\midrule
SL & 0.754 & 0.660 & 0.701 & 0.759 & 0.670 & 0.705 & 0.757 & 0.669 & 0.696 \\
FL & 0.737 & 0.653 & 0.656 & 0.737 & 0.651 & 0.661 & 0.765 & 0.633 & 0.694 \\
\bottomrule
\end{tabular}
\vspace{-15pt}
\end{table}
\subsection{Compute and Memory Cost: SL vs FL}
\label{sec:fed}
The FL baseline trains the same trunk-LoRA adapters locally and aggregates them with sample-weighted FedAvg, using the same $2{,}000$-sample, $10$-client setup. SL exceeds FL in average ANLS ($0.708$ vs.\ $0.687$) and wins in eight of nine settings (mean advantage $+0.021$; Table~\ref{tab:sl_fl}). The methods place cost differently (Table~\ref{tab:params}): each SL client trains only $3.68$M parameters ($0.114\%$ of the $3.21$B model) at $\sim$$0.08$ Tera FLOating-Point operations per step (TFLOPs) and $\sim$$1.3$~GB, while the heavy trunk ($\sim$$158$~TFLOPs, $15$--$25$~GB) runs once server-side. In FL, every client instead runs the full model ($\sim$$158$~TFLOPs, $15$--$25$~GB). SL thus keeps clients lightweight at roughly three orders of magnitude lower per-client compute.
\begin{table}[htp]
\centering
\caption{Per-side compute and memory costs in our setup.}
\label{tab:params}
\vspace{-5pt}
\renewcommand{\arraystretch}{1.2}
\setlength{\tabcolsep}{3pt}
\begin{tabular}{lcccc}
\toprule
Method / side & \makecell{Frozen\\base} & \makecell{Trainable\\params} & \makecell{Compute\\/ step} & \makecell{VRAM\\(train)} \\
\midrule
SL client      & $\sim$0.39B & 3.68M  & $\sim$0.08~TFLOPs & $\sim$1.3~GB \\
SL server      & $\sim$3.21B & 24.31M & $\sim$158~TFLOPs  & 15--25~GB \\
FL client (ea.)& $\sim$3.21B & 24.31M & $\sim$158~TFLOPs  & 15--25~GB \\
FL server      & 0           & 0      & $\sim$0           & $\sim$0 (CPU) \\
\bottomrule
\end{tabular}
\vspace{-15pt}
\end{table}

\subsection{The Generalist Model}
\label{sec:gen}
\begin{figure}[t]
\centering
\includegraphics[width=\columnwidth]{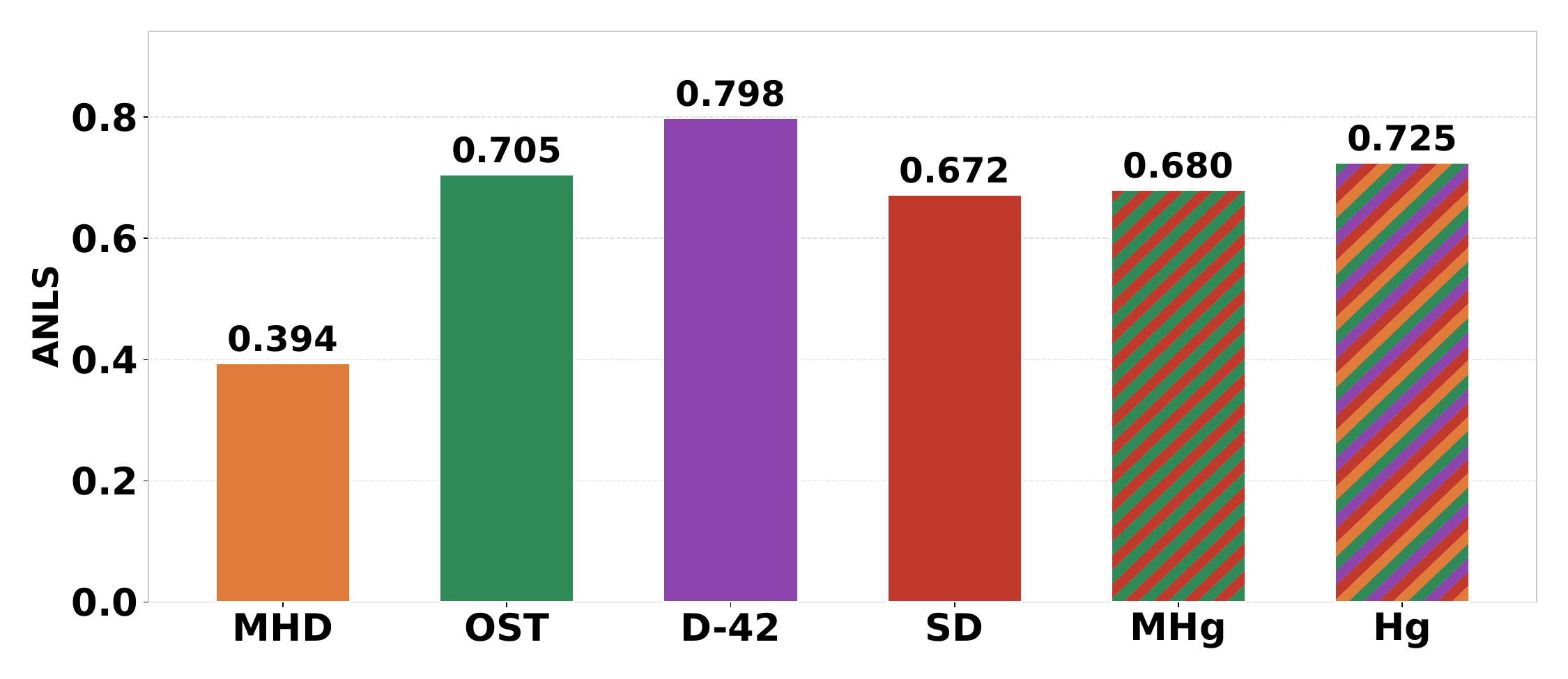}
\vspace{-20pt}
\caption{Generalist ANLS across test datasets.}
\vspace{-10pt}
\label{fig:gen}
\end{figure}

\begin{figure}[t]
\begin{subfigure}{0.49\columnwidth}
\centering
\includegraphics[width=\linewidth]{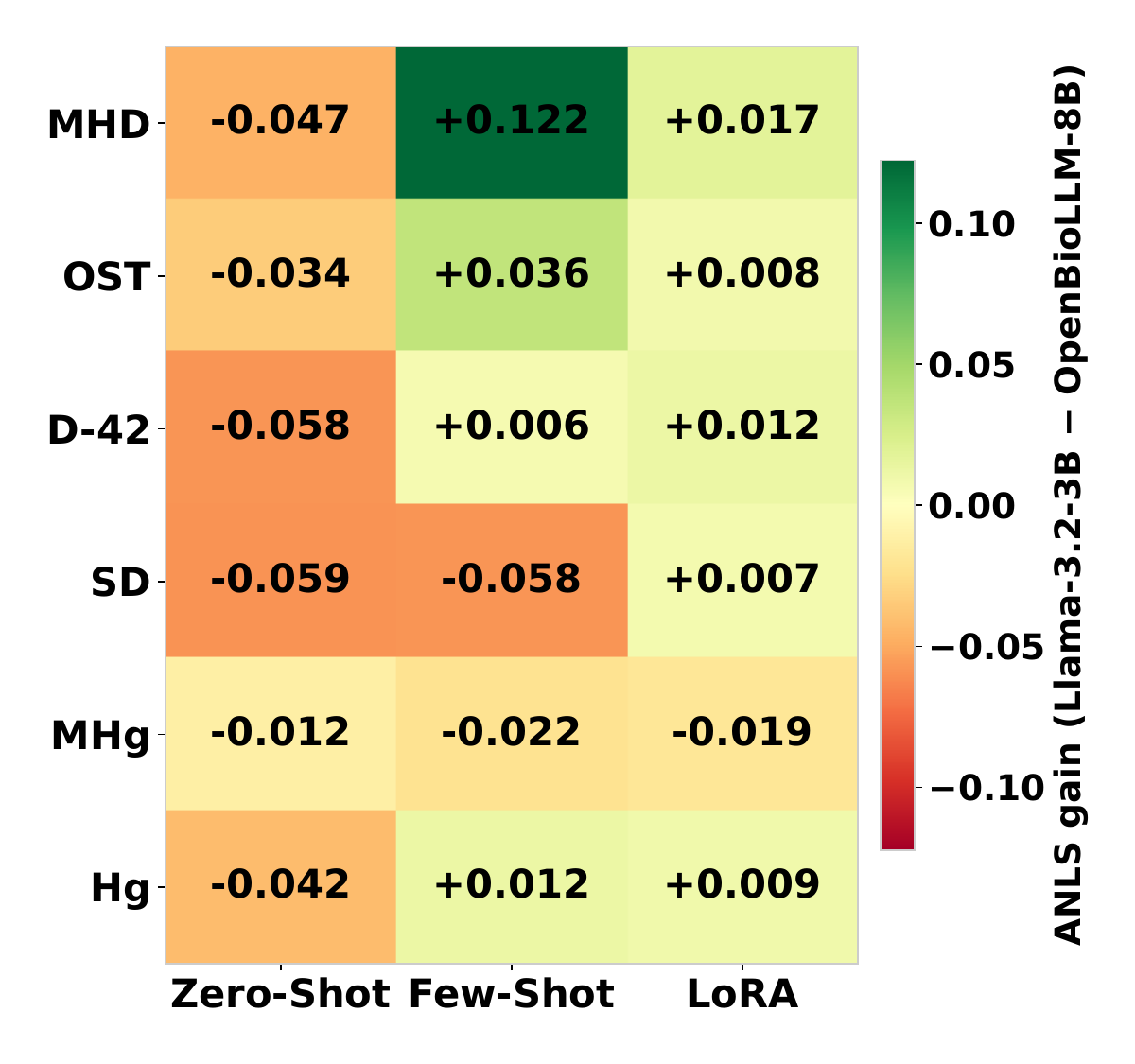}
\vspace{-20pt}
\caption{}
\label{fig:backbonegain}
\end{subfigure}
\hfill
\begin{subfigure}{0.49\columnwidth}
\centering
\includegraphics[width=\linewidth]{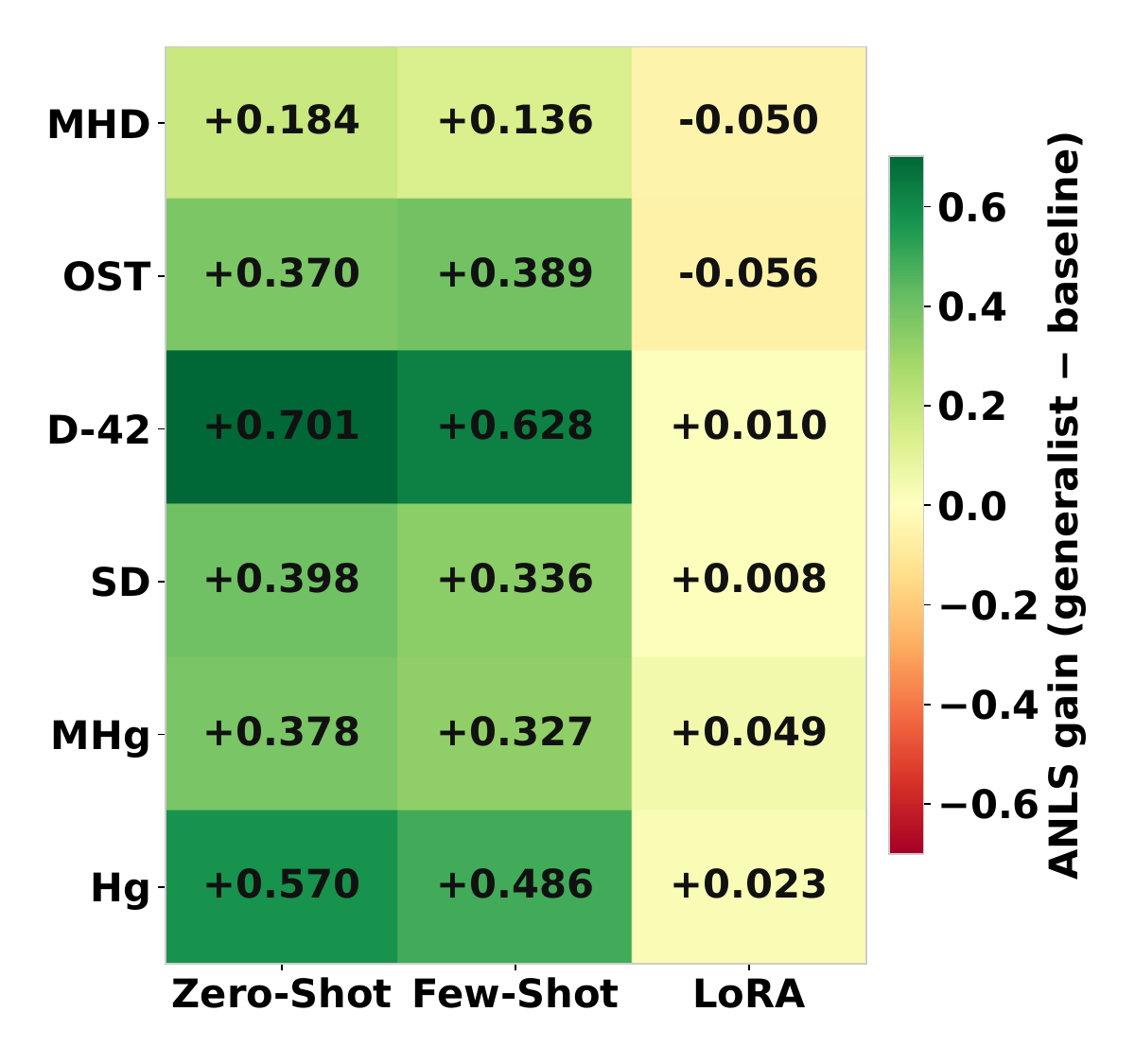}
\vspace{-20pt}
\caption{}
\label{fig:gengain}
\end{subfigure}
\vspace{-15pt}
\caption{ANLS gain heatmaps: (a) OpenBioLLM vs.\ LLaMA; (b) Generalist vs.\ baselines on LLaMA.}
\vspace{-15pt}
\label{fig:gainheatmaps}
\end{figure}

The \textit{generalist model} is trained on the merged collection of all heterogeneous datasets in the symmetric-volume mode and evaluated on all six held-out test sets (Fig.~\ref{fig:gen}). The training pool has $10{,}000$ samples spread evenly across $10$ clients ($1{,}000$ each), a fivefold increase over the grid. It averages $0.662$ ANLS, transfers well to D-42 ($0.798$) and Hg ($0.725$), and performs worst on the feature-poor MHD set ($0.394$). This increase gives only a small gain over the matching grid setting, suggesting the model already generalizes from limited data. Fig.~\ref{fig:gengain} compares it with per-dataset baselines: it beats zero- and few-shot on every dataset, by up to $+0.701$ ANLS over zero-shot on D-42, and remains competitive with dataset-specific LoRA. This shows it captures transferable structure rather than one fixed schema.

\subsection{Performance under FT-Transformer Baseline}
\label{sec:ft}
\begin{figure}[t]
\centering
\includegraphics[width=\columnwidth]{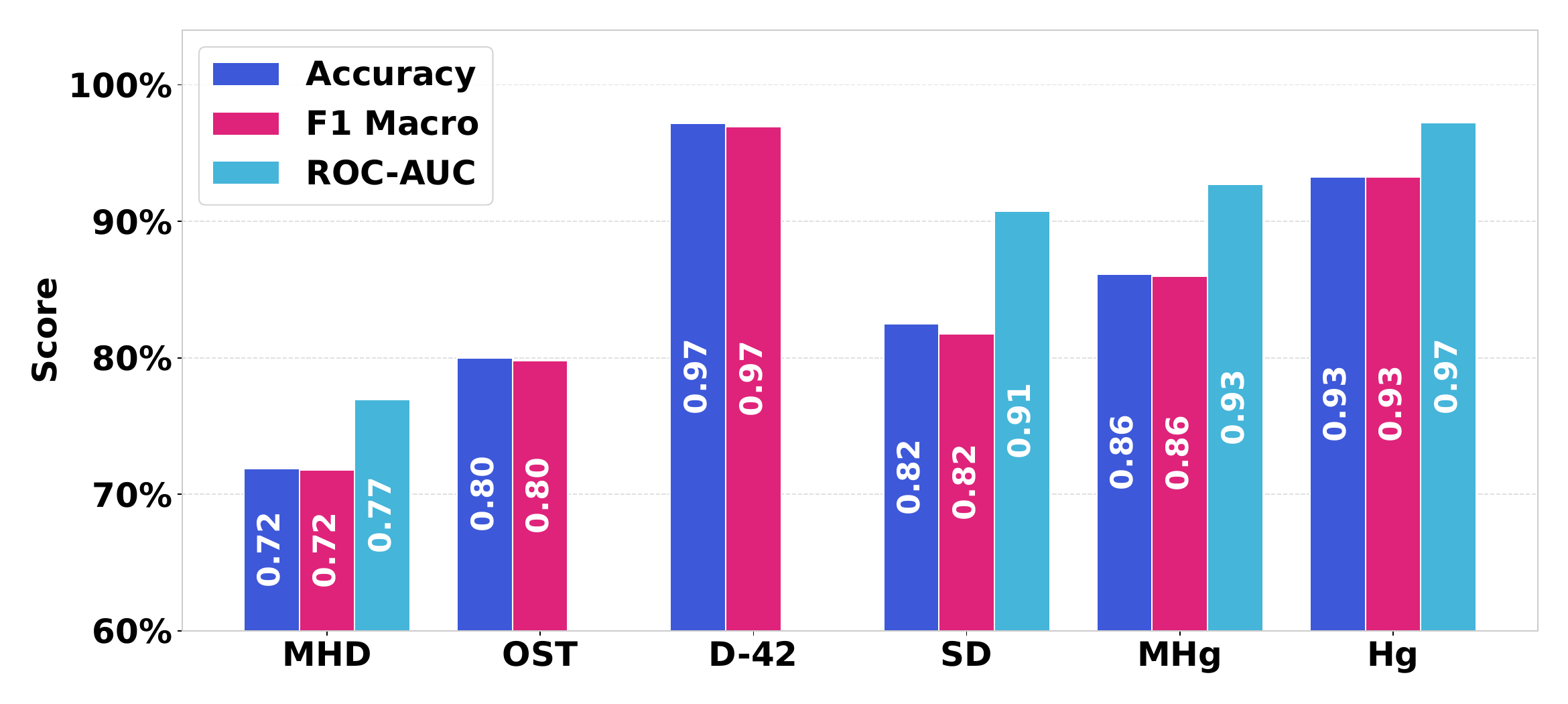}
\vspace{-20pt}
\caption{FT-Transformer results across datasets.}
\vspace{-15pt}
\label{fig:ft}
\end{figure}

We evaluate a fully supervised, non-LLM FT-Transformer as a per-dataset upper bound (Fig.~\ref{fig:ft}). It averages $0.852$ accuracy, peaks on D-42 ($0.972$) and Hg ($0.932$), and is lowest on the feature-poor MHD set ($0.719$); ROC-AUC is shown only for binary targets. The model is accurate but schema-bound: it depends on a fixed feature set and cannot handle a new survey without retraining, the heterogeneity our LLM-based framework is designed to absorb.

\subsection{Research Question Discussion and Limitations}
\label{sec:discussion}
For \textbf{RQ1}, the framework beats prompting baselines and matches or exceeds FL ($0.708$ vs.\ $0.687$) even when clients hold different schemas, confirming that schema-aware serialization enables privacy-preserving cross-schema sharing. For \textbf{RQ2}, performance is robust: it varies by under $0.01$ ANLS across volume levels, drops slightly with heterogeneity, and peaks at $+0.045$ over FL in symmetric/high setting. For \textbf{RQ3}, SL keeps client lightweight ($3.68$M parameters, $\sim$$1.3$~GB, $\sim$$0.08$~TFLOPs) against FL's full-model client ($15$--$25$~GB, $\sim$$158$~TFLOPs, $2000$ times heavier), and the OpenBioLLM-8B and cross-dataset results ($0.662$ average) show gains hold across backbones and unseen data. The evaluation covers four survey domains with $500$-sample test sets and order-of-magnitude compute estimates. On privacy, only cut-layer activations and gradients leave each institution; we do not claim these tensors resist reconstruction, and defending against model-inversion attacks is outside our scope.

\subsection{Future Work}
\label{sec:future_work}

Future work will focus on stronger privacy guarantees, broader validation, and practical deployment. This includes protecting cut-layer representations with differential privacy and secure aggregation, evaluating larger and more diverse clinical datasets, and measuring computation, communication, and runtime costs. Further studies should examine domain-specific models, uncertainty, fairness, and clinician-in-the-loop validation. Multi-institution deployment will also be important for testing the framework under real governance and infrastructure constraints.

%% file: Sections/conclusion.tex
\section{Conclusion}
\label{sec:conclusion}
We presented a privacy-preserving SL framework, with a LoRA-adapted LLM as a collaborative encoder, that predicts mental distress from survey data, letting institutions share predictive knowledge while the responses remain local. Central to the design is a schema-aware serialization that maps heterogeneous records into a shared representation. Across a $3\times3$ grid of volume imbalance and schema heterogeneity, it averaged $0.708$ ANLS, stayed stable under increasing heterogeneity, and beat a federated baseline in eight of nine settings while keeping per-client compute three orders of magnitude lighter. It also surpassed zero- and few-shot prompting, stayed competitive with centralized LoRA through a single shared model, generalized across LLM families, and remained effective from a small training pool.

\section{Acknowledgement}
The authors acknowledge the National Artificial Intelligence Research Resource (NAIRR) Pilot for contributing to this research result with  NCSA Delta GPU access (NAIRR260054).

%% file: References.bib
@book{world2022world,
  title={World mental health report: Transforming mental health for all},
  author={World Health Organization},
  year={2022},
  publisher={World Health Organization}
}

@article{lovibond1995structure,
  title={The structure of negative emotional states: Comparison of the Depression Anxiety Stress Scales (DASS) with the Beck Depression and Anxiety Inventories},
  author={Lovibond, Peter F and Lovibond, Sydney H},
  journal={Behaviour research and therapy},
  volume={33},
  number={3},
  pages={335--343},
  year={1995},
  publisher={Elsevier}
}

@inproceedings{hegselmann2023tabllm,
  title={Tabllm: Few-shot classification of tabular data with large language models},
  author={Hegselmann, Stefan and Buendia, Alejandro and Lang, Hunter and Agrawal, Monica and Jiang, Xiaoyi and Sontag, David},
  booktitle={International conference on artificial intelligence and statistics},
  pages={5549--5581},
  year={2023},
  organization={PMLR}
}

@article{dinh2022lift,
  title={Lift: Language-interfaced fine-tuning for non-language machine learning tasks},
  author={Dinh, Tuan and Zeng, Yuchen and Zhang, Ruisu and Lin, Ziqian and Gira, Michael and Rajput, Shashank and Sohn, Jy-yong and Papailiopoulos, Dimitris and Lee, Kangwook},
  journal={Advances in Neural Information Processing Systems},
  volume={35},
  pages={11763--11784},
  year={2022}
}

@inproceedings{yan2024making,
  title={Making pre-trained language models great on tabular prediction},
  author={Yan, Jiahuan and Zheng, Bo and Xu, Hongxia and Zhu, Yiheng and Chen, Danny and Sun, Jimeng and Wu, Jian and Chen, Jintai},
  booktitle={International Conference on Learning Representations},
  year={2024}
}

@article{hu2022lora,
  title={Lora: Low-rank adaptation of large language models.},
  author={Hu, Edward J and Shen, Yelong and Wallis, Phillip and Allen-Zhu, Zeyuan and Li, Yuanzhi and Wang, Shean and Wang, Liang and Chen, Weizhu and others},
  journal={Iclr},
  volume={1},
  number={2},
  pages={3},
  year={2022}
}

@article{gaber2025federated,
  title={Federated Learning Meets LLMs: Feature Extraction From Heterogeneous Clients},
  author={Gaber, Abdelrhman and Abd-Eltawab, Hassan and Abuzied, Youssif and ElMahdy, Muhammad and ElBatt, Tamer},
  journal={arXiv preprint arXiv:2510.00065},
  year={2025}
}

@article{lin2024splitlora,
  title={Splitlora: A split parameter-efficient fine-tuning framework for large language models},
  author={Lin, Zheng and Hu, Xuanjie and Zhang, Yuxin and Chen, Zhe and Fang, Zihan and Chen, Xianhao and Li, Ang and Vepakomma, Praneeth and Gao, Yue},
  journal={arXiv preprint arXiv:2407.00952},
  year={2024}
}

@article{lin2026hsplitlora,
  title={HSplitLoRA: A heterogeneous split parameter-efficient fine-tuning framework for large language models},
  author={Lin, Zheng and Zhang, Yuxin and Chen, Zhe and Fang, Zihan and Chen, Xianhao and Vepakomma, Praneeth and Ni, Wei and Luo, Jun and Gao, Yue},
  journal={IEEE Transactions on Mobile Computing},
  year={2026},
  publisher={IEEE}
}

@article{vaswani2017attention,
  title={Attention is all you need},
  author={Vaswani, Ashish and Shazeer, Noam and Parmar, Niki and Uszkoreit, Jakob and Jones, Llion and Gomez, Aidan N and Kaiser, {\L}ukasz and Polosukhin, Illia},
  journal={Advances in neural information processing systems},
  volume={30},
  year={2017}
}

@article{grattafiori2024llama,
  title={The llama 3 herd of models},
  author={Grattafiori, Aaron and Dubey, Abhimanyu and Jauhri, Abhinav and Pandey, Abhinav and Kadian, Abhishek and Al-Dahle, Ahmad and Letman, Aiesha and Mathur, Akhil and Schelten, Alan and Vaughan, Alex and others},
  journal={arXiv preprint arXiv:2407.21783},
  year={2024}
}

@misc{OpenBioLLMs,
  author = {Ankit Pal, Malaikannan Sankarasubbu},
  title = {OpenBioLLMs: Advancing Open-Source Large Language Models for Healthcare and Life Sciences},
  year = {2024},
  publisher = {Hugging Face},
  journal = {Hugging Face repository},
  howpublished = {\url{https://huggingface.co/aaditya/OpenBioLLM-Llama3-70B}}
}

@article{vepakomma2018split,
  title={Split learning for health: Distributed deep learning without sharing raw patient data},
  author={Vepakomma, Praneeth and Gupta, Otkrist and Swedish, Tristan and Raskar, Ramesh},
  journal={arXiv preprint arXiv:1812.00564},
  year={2018}
}

@inproceedings{thapa2022splitfed,
  title={Splitfed: When federated learning meets split learning},
  author={Thapa, Chandra and Arachchige, Pathum Chamikara Mahawaga and Camtepe, Seyit and Sun, Lichao},
  booktitle={Proceedings of the AAAI conference on artificial intelligence},
  volume={36},
  number={8},
  year={2022}
}

@article{gorishniy2021revisiting,
  title={Revisiting deep learning models for tabular data},
  author={Gorishniy, Yury and Rubachev, Ivan and Khrulkov, Valentin and Babenko, Artem},
  journal={Advances in neural information processing systems},
  volume={34},
  pages={18932--18943},
  year={2021}
}

@inproceedings{mcmahan2017communication,
  title={Communication-efficient learning of deep networks from decentralized data},
  author={McMahan, Brendan and Moore, Eider and Ramage, Daniel and Hampson, Seth and y Arcas, Blaise Aguera},
  booktitle={Artificial intelligence and statistics},
  year={2017},
  organization={Pmlr}
}

@misc{osmi2014,
  author       = {{Open Sourcing Mental Illness (OSMI)}},
  title        = {Mental Health in Tech Survey},
  year         = {2014--2023},
  howpublished = {\url{https://osmihelp.org/research}},
  note         = {Annual survey data collected 2014--2023. Accessed 2024.}
}

@misc{dass42_openpsychometrics,
  author       = {{Open-Source Psychometrics Project}},
  title        = {{DASS-42} Dataset: Depression Anxiety Stress Scales Online Administration},
  year         = {2019},
  howpublished = {\url{https://openpsychometrics.org/_rawdata/}},
  note         = {Data collected 2017--2019. Accessed 2024.}
}

@misc{student_depression_kaggle,
  author       = {Shodolamu Opeyemi},
  title        = {Student Depression Dataset},
  year         = {2024},
  month        = nov,
  howpublished = {\url{https://www.kaggle.com/datasets/hopesb/student-depression-dataset}},
  note         = {Kaggle dataset, last accessed August 2026}
}

@misc{mental_health_general_kaggle,
  author       = {Bhavik Jikadara},
  title        = {Mental Health Dataset},
  year         = {2023},
  howpublished = {\url{https://www.kaggle.com/datasets/bhavikjikadara/mental-health-dataset}},
  note         = {Kaggle dataset, last accessed August 2026}
}

@article{cuijpers2023world,
  title={The WHO world mental health report: a call for action},
  author={Cuijpers, Pim and Javed, Afzal and Bhui, Kamaldeep},
  journal={The British Journal of Psychiatry},
  volume={222},
  year={2023},
  publisher={Cambridge University Press}
}

@article{thornicroft2022lancet,
  title={The Lancet Commission on ending stigma and discrimination in mental health},
  author={Thornicroft, Graham and Sunkel, Charlene and Aliev, Akmal Alikhon and Baker, Sue and Brohan, Elaine and El Chammay, Rabih and Davies, Kelly and Demissie, Mekdes and Duncan, Joshua and Fekadu, Wubalem and others},
  journal={The Lancet},
  volume={400},
  number={10361},
  pages={1438--1480},
  year={2022},
  publisher={Elsevier}
}

@inproceedings{ye2024towards,
  title={Towards cross-table masked pretraining for web data mining},
  author={Ye, Chao and Lu, Guoshan and Wang, Haobo and Li, Liyao and Wu, Sai and Chen, Gang and Zhao, Junbo},
  booktitle={Proceedings of the ACM Web Conference 2024},
  pages={4449--4459},
  year={2024}
}

@misc{meta2024llama32,
  title        = {Llama 3.2: Revolutionizing Edge {AI} and Vision with Open, Customizable Models},
  author       = {{Meta AI}},
  year         = {2024},
  howpublished = {\url{https://huggingface.co/meta-llama/Llama-3.2-3B-Instruct}},
  note         = {Llama-3.2-3B-Instruct model card; released September 25, 2024}
}

@article{shatte2019machine,
  title={Machine learning in mental health: a scoping review of methods and applications},
  author={Shatte, Adrian BR and Hutchinson, Delyse M and Teague, Samantha J},
  journal={Psychological medicine},
  volume={49},
  number={9},
  pages={1426--1448},
  year={2019},
  publisher={Cambridge University Press}
}

@article{hipaa1996,
  title={Health insurance portability and accountability act of 1996},
  author={Act, Accountability and others},
  journal={Public law},
  volume={104},
  number={191},
  pages={1--16},
  year={1996}
}

@techreport{samhsa2023nsduh,
  title        = {Key Substance Use and Mental Health Indicators in the United States: Results from the 2022 National Survey on Drug Use and Health},
  author       = {{Substance Abuse and Mental Health Services Administration}},
  year         = {2023},
  institution  = {Center for Behavioral Health Statistics and Quality, Substance Abuse and Mental Health Services Administration},
  number       = {HHS Publication No. PEP23-07-01-006, NSDUH Series H-58},
  address      = {Rockville, MD}
}

@article{voigt2017eu,
  title={The eu general data protection regulation (gdpr)},
  author={Voigt, Paul and Von dem Bussche, Axel},
  journal={A practical guide, 1st ed., Cham: Springer International Publishing},
  volume={10},
  number={3152676},
  pages={10--5555},
  year={2017},
  publisher={Springer}
}

@article{khalil2024federated,
  title={Federated learning for privacy-preserving depression detection with multilingual language models in social media posts},
  author={Khalil, Samar Samir and Tawfik, Noha S and Spruit, Marco},
  journal={Patterns},
  volume={5},
  number={7},
  year={2024},
  publisher={Elsevier}
}

@article{gao2022survey,
  title={A survey on heterogeneous federated learning},
  author={Gao, Dashan and Yao, Xin and Yang, Qiang},
  journal={arXiv preprint arXiv:2210.04505},
  year={2022}
}
